\documentclass[runningheads]{llncs}

\usepackage[sort]{cite}

\usepackage{graphicx}

\usepackage{xcolor}

\definecolor{darkblue}{rgb}{0,0,0.75}
\usepackage[hidelinks,pdftex,pdftitle={Title},pdfauthor={Author}, colorlinks=false, bookmarksnumbered, citecolor=darkblue, urlcolor=darkblue, linkcolor=darkblue, pdfpagemode=UseNone]{hyperref}

\usepackage{pgfplots}
\pgfplotsset{width=10cm,compat=1.9}
\usepgfplotslibrary{external}
\usepackage{subcaption, booktabs}

\usepackage{amsmath}
\usepackage{amssymb}
\usepackage{mathtools}
\usepackage{makecell}
\usepackage{multirow}
\usepackage{appendix}
\usepackage[misc]{ifsym}

\DeclareMathOperator*{\argmax}{arg\,max}

\DeclareMathOperator*{\xref}{X_{\mathop{ref}}}

\definecolor{lamarrBlue}{rgb}{0.10980392, 0.26666667, 0.43529412}
\definecolor{lamarrGreen}{rgb}{0.2555, 0.47, 0.25}
\definecolor{lamarrOrange}{rgb}{0.97, 0.6, 0.03}
\definecolor{lamarrRed}{rgb}{0.74509804, 0.40392157, 0.47843137} 

\newcommand\blfootnote[1]{%
  \begingroup
  \renewcommand\thefootnote{}\footnote{#1}%
  \addtocounter{footnote}{-1}%
  \endgroup
}

\begin{document}

\title{An Empirical Evaluation of the Rashomon Effect in Explainable Machine Learning}

\titlerunning{Empirical Evaluation of the Rashomon Effect}
\author{Sebastian Müller (\href{mailto:semueller@uni-bonn.de}{\Letter})\inst{1, 4}\orcidID{0000-0002-0778-9695} \and
Vanessa Toborek\inst{1,4}\orcidID{0009-0009-8372-8251}\and
Katharina Beckh\inst{3,4}\orcidID{0000-0002-7824-6647}\and
Matthias Jakobs\inst{2,4}\orcidID{0000-0003-4607-8957} \and
Christian Bauckhage\inst{1,3,4}\orcidID{0000-0001-6615-2128}\and
Pascal Welke\inst{5}\orcidID{0000-0002-2123-3781}
}
\authorrunning{Müller et al.}
\institute{University of Bonn, Bonn, Germany \and
TU Dortmund University, Dortmund, Germany \and
Fraunhofer IAIS, Sankt Augustin, Germany \and
Lamarr Institute, Germany \and
TU Wien, Vienna, Austria \\
}

\maketitle              

\blfootnote{Accepted for presentation at European Conference on Machine Learning and Principles and Practice of Knowledge Discovery in Databases (ECML/PKDD 2023)}
\setcounter{footnote}{0}

\begin{abstract}
The Rashomon Effect describes the following phenomenon: for a given dataset there may exist many models with equally good performance but with different solution strategies. 
The Rashomon Effect has implications for Explainable Machine Learning, especially for the comparability of explanations.
We provide a unified view on three different comparison scenarios and conduct a quantitative evaluation across different datasets, models, attribution methods, and metrics.
We find that hyperparameter-tuning plays a role and that metric selection matters.  
Our results provide empirical support for previously anecdotal evidence and exhibit challenges for both scientists and practitioners.

\end{abstract}

\section{Introduction}

We demonstrate the impact of the Rashomon Effect when analyzing ML models.
The Rashomon Effect \cite{breiman2001twocultures} describes the phenomenon that there may exist many models within a hypothesis class which solve a dataset equally well. 
The set of these models is referred to as the Rashomon Set \cite{fisher2019all,xin2022exploring}.
From a data-centric perspective this phenomenon is also called Predictive Multiplicity \cite{marx2020predictive}, meaning that there exist many strategies to solve a task on a dataset.
Other works use Rashomon Sets to analyze and describe data \cite{fisher2019all, semenova2022existence}.
Somewhat surprisingly, the Rashomon Effect has not yet found wider attention in the Explainable Machine Learning (XML) literature. 
Although a few works have observed the effect it was only anecdotally or without referring to its proper name \cite{watson2022agree,guidotti2018assessing,leventi2022rashomon}.

XML has recently become a very active area of research and numerous explanation methods exist \cite{adadi2018peeking,burkart2021survey,molnar2022}. 
Many approaches explain black-box models in a post-hoc manner by providing attribution scores \cite{lundbergUnifiedApproachInterpreting2017a,ribeiroWhyShouldTrust2016} which assign each input dimension a numerical value that represents this feature's importance with respect to the model decision. 
Attribution scores are used to answer questions such as ``What feature was the most important in this input sample?'' and have been used to uncover spurious correlations in the data \cite{schramowski2020making} and biased behavior of models \cite{liu-avci-2019-incorporating}.
However, attribution scores are sometimes ambiguous and their interpretation depends on the application context.
It is hard to decide at what magnitude a feature is still important, particularly, if magnitudes of attribution scores can be sorted into an evenly descending order.
It follows that the task of comparing different attribution methods is a difficult problem.
Several works touch upon the problem of explanation comparison \cite{bogun22saliency,atanasova2020diagnostic,neely2021order,krishnaDisagreementProblemExplainable2022,watson2022agree} from different perspectives.

Our main contribution is an empirical analysis of one novel and two existing perspectives, 1) demonstrating model-specific sensitivity regarding the hyperparameter choice for explanation methods, 2) comparison of different explanations from the same attribution method on differently initialized but otherwise identical model architectures \cite{watson2022agree,atanasova2020diagnostic} and  3) the disagreement between different explanations applied to the same architecture and parameterization \cite{krishnaDisagreementProblemExplainable2022,neely2021order}.
We place these three perspectives into a unified framework to investigate how the Rashomon Effect manifests itself in each situation. 
Our evaluation is conducted on four datasets of entirely different nature, analyzing differences in models explained by five popular attribution methods using both naive and established human-centered similarity measures.\footnote{Our code is available at \href{https://github.com/lamarr-xai-group/RashomonEffect}{github.com/lamarr-xai-group/RashomonEffect}}
Our results highlight the need to fine-tune the hyperparameters of XML methods on a per-model basis.
We do find empirical support for the disagreement problem, meaning practitioners cannot expect consistent explanations across methods. 
Further, the high solution diversity across models hinders the use of XML as an epistemic tool.

Next, Section \ref{sec:2} discusses how we connect different parts of the literature for our analysis. Section \ref{sec:methods} describes the experimental setup in detail. Sections \ref{sec:111}, \ref{sec:011} and \ref{sec:110} present results and discuss the three perspectives we analyze. Section \ref{sec:4} summarizes our main findings. Section \ref{sec:5} concludes.

\section{Comparing Attribution Scores}
\label{sec:2}
Given a classifier and a datum, an attribution method assigns each input dimension a numerical value that represents this feature's importance with respect to the model decision.
Hence, an attribution scoring depends on three variables: 1) the model, 2) the input sample, and 3) the attribution method.
This distinction enables us to systematically investigate the consequences of the Rashomon Effect on established and novel perspectives in XML in one framework. 
This framework is the first to bring the different perspectives into a unifying picture which we present in Table \ref{tab:discussion}. 
Our main mode of comparison is centered around comparing pairs of attribution scores. Hence, we assume that the scores belong to the same sample from the same dataset. We investigate model- or attribution method-dependent effects and do not consider the scenario where the data is the same but both models and methods are different.

\textbf{Numerical Stability} (111): In Section \ref{sec:111} we discuss the scenario where the same model and same attribution method are applied to the same sample. 
This perspective is relevant to non-deterministic explanation methods that can be controlled by hyperparameters. We investigate whether there are model-specific differences regarding optimal parameter choice and find that the hyperparameter choice is significantly dependent on both the investigated model and dataset.
This suggests that blindly applying non-optimal hyperparameters can lead to erroneous explanations and thus wrong takeaways in an application scenario. 
This need for rigorous hyperparameter tuning is mostly overlooked in the literature.

\begin{table}[t]
    \caption{
    We investigate the Rashomon Effect in Explainable Machine Learning for a set of models and a set of attribution methods.
    Three interesting scenarios arise for a fixed input-sample from a given dataset.}
    \centering
    \begin{tabular}{b{1.3cm}b{1.3cm}b{1.5cm}b{4.5cm}b{1.8cm}}
    \makecell{Same \\ Model} & \makecell{Same \\ Sample} & \makecell{Same Attr \\ Method} & \makecell{Scenario} & \makecell{Examples} \\ 
    \midrule[1px]
    \makecell{1} & \makecell{1} & \makecell{1} & \makecell{Numerical Stability} &  \makecell{$-$} \\ 
    \midrule    
    \makecell{0} & \makecell{1} & \makecell{1} & \makecell{Solution Diversity} & \makecell{\cite{fisher2019all}  \cite{hooker2019benchmark} \cite{watson2022agree} } \\ \midrule
    \makecell{1} & \makecell{1} & \makecell{0} & \makecell{Disagreement Problem} & \makecell{\cite{elshawi2021interpretability}\cite{krishnaDisagreementProblemExplainable2022} \cite{neely2021order}  } \\ 
\bottomrule
    \end{tabular}
    \label{tab:discussion}
\end{table}

\textbf{Solution Diversity} (011): We can compare how similar or dissimilar two models are w.r.t. their solution strategies by comparing explanations that were computed for each of them using the same attribution method. Comparing any two models not only by one, but by the average difference on explanations over several samples, will only be able to measure a difference, if two models consistently behave differently.
This is a coarse, but sufficiently sensitive measure.
Using this measure as a basis, we provide a large quantitative view of the Rashomon Effect itself, recently also observed in \cite{watson2022agree}. In Section \ref{sec:011}, we extend existing results by comparing substantially more models on additional data domains and investigate how diverse the strategies of the models within a Rashomon Set are. We observe very high diversity in most cases and discuss practical implications for machine learning (ML) as an epistemic tool \cite{zednik2022scientific,roscher2020explainable}.

\textbf{Disagreement Problem} (110): Aiming to find the ``right'' explanation, prior work compared different attribution methods applied to the same model on the same sample. It was found that explanations of different attribution methods often differ significantly, which is now known as the Disagreement Problem \cite{krishnaDisagreementProblemExplainable2022,hancoxli2020,neely2021order,atanasova2020diagnostic,flora2022comparing,elshawi2021interpretability,han2022explanation}.
So far, the Disagreement Problem was only reported on individual or a very small number of models. It has not been sufficiently explored whether the disagreement actually is model-dependent, i.e., whether any pair of attribution methods is consistently less similar than other pairs across models. We investigate this question in Section \ref{sec:110}. We provide quantitative support for anecdotal observations from the literature and add practically relevant insights.

A fundamental question is which metric should be used to compare two attribution scores. One possible approach is to use feature (dis-)agreement, i.e. the overlap of  the top-k ``most important'' features, which ML practitioners indicated as a key measure for disagreement \cite{krishnaDisagreementProblemExplainable2022}. Along the same lines, ranking correlation measures are used, such as Kendall's $\tau$ \cite{neely2021order}.
Another option is to base the comparison on typical distance measures, such as cosine similarity \cite{bogun22saliency} or Euclidean distance \cite{elshawi2021interpretability,mucke2022check}. 
In previous studies, only one metric or metric type has been considered. In this work, we provide a comparison of both Euclidean and (dis-)agreement based measures.

\section{Experimental Framework}
\label{sec:methods}
Before we report our results we introduce the experimental setup. 
To emphasize the extent of the Rashomon Effect we will remove randomness from the training process with the exception of model initialization.

\subsection{Datasets} 
\label{sec:data}
For the comparison we chose four publicly available datasets. \textit{AG News} \cite{zhangCharacterlevelConvolutionalNetworks2016}, a benchmark dataset for text classification with an average sentence length of 43 words. 
Three tabular datasets containing only real valued variables: \textit{Dry Bean} \cite{data_beans}, a 16-dimensional multi-class dataset with 7 classes of dry beans, \textit{Breast Cancer Wisconsin (Diagnostic)} \cite{data_breastcancer}, a classical dataset posing a binary classification problem over 30 features, and \textit{Ionosphere} \cite{data_ionosphere}, a binary classification problem over 34 features based on radar signal returns. The amount of data available with each dataset differs greatly. A random subset $\xref$ was held out from each dataset during training and later used for the computation of explanations. $\xref$ contains 300 samples for AG News, 1050 for Dry Bean, 114 for Breast Cancer and 71 for Ionosphere.

\subsection{Models: Architecture, Training and Selection}
\label{sec:models}
For the tabular datasets we use small, fully connected Feed-Forward Neural Networks with ReLU activation functions. Models for Dry Bean, Breast Cancer and Ionosphere use 3x16, 16 and 8 neurons, respectively.
For the AG News dataset we use a Bi-LSTM model with 128 dimensions for each direction and a fully connected output layer. We learn a 128 dimensional word embedding from scratch. We use the softmax function as output activation in all models.

We trained 100 models on each tabular dataset and 20 models on AG News. We fixed all random aspects of the model training except for the initialization of the network parameters. Each model observed exactly the same amount of data in the exact same order. All differences in model behavior will thus only stem from the initialization.
To build the final Rashomon Set for each dataset, we choose all models with at most $5\%$ difference in accuracy to the best model. 
With the exception of the Ionosphere dataset, nearly all models are selected. We present average model accuracy and average pairwise output similarity computed with the Jensen-Shannon-Distance over $\xref$ in Table~\ref{tab:comparison}. All models achieve a high accuracy and are nearly indistinguishable by their output distributions.

\begin{table}[t]
    \caption{Mean accuracy and mean pairwise Jensen-Shannon-Distance (JSD) of all models over $\xref$. All models were selected to lie within 5\% accuracy of the best model. According to both metrics, all models perform nearly indistinguishably. JSD is bounded to $[0,1]$.}
    \centering
    \begin{tabular*}{\textwidth}{@{\extracolsep{\fill}}lcccc@{}}
        \toprule
        & AG News & Dry Bean & Breast Cancer & Ionosphere \\ 
        \midrule
        \makecell[l]{Mean accuracy on $\xref$} & $0.91 \pm 0.01 $ & $0.89 \pm 0.01 $ & $0.95 \pm 0.01 $ & $0.86 \pm 0.01 $\\ 
        \makecell[l]{Mean JSD on $\xref$} & $0.0315 \pm 0.004$ & $0.0207 \pm 0.004$ & $0.0019 \pm 0.001$ & $0.0103 \pm 0.007$ \\
        \bottomrule \\
    \end{tabular*}
    \label{tab:comparison} \end{table}

\subsection{Attribution Methods}
\label{sec:attrmeth}
We compare five attribution methods. From the family of gradient based methods we use Vanilla Grad (VG) \cite{simonyan2013deep}, Smooth Grad (SG) \cite{smilkov2017smoothgrad}, and Integrated Gradient (IG) \cite{sundararajan2017axiomatic}.
From the family of perturbation based methods we include KernelSHAP (KS) \cite{lundbergUnifiedApproachInterpreting2017a} and LIME (LI) \cite{ribeiroWhyShouldTrust2016} for which we use the implementations provided by Captum\footnote{See project page at \href{https://github.com/pytorch/captum}{github.com/pytorch/captum}}. For IG, KS, and LI we use zero-baselines. SG samples with a noise ratio of 10\%.
Hyperparameters that further impact approximation behavior will be discussed in Section \ref{sec:111}.

\subsection{Model Dissimilarity Measures based on Attribution Scores}

We use the following formula to express the scenarios in Table~\ref{tab:discussion}:

\begin{equation}
    \mathop{\mathcal{D}}(f_a, f_b, X, \phi_1, \phi_2, d) = \frac{1}{|X|} \sum_{x \in X} d(\phi_1(f_a, x), \phi_2(f_b, x))
    \label{eq:base}
\end{equation}

where $f_a,\ f_b \in R$ are classifier functions from our Rashomon Set, $X \subseteq \xref: \{ x | x \in \xref \land \argmax{f_a(x)} = \argmax{f_b(x)} \}$ is a subset of the reference set where both classifiers agree on the label, $\phi_1,\phi_2 \in \Phi = \{$VG, SG, IG, KS, LI$\}$ are the aforementioned attribution methods and
$d \in D = \{${\color{lamarrBlue}{Feature Disagreement}}, {\color{lamarrGreen}{Sign Disagreement}}, {\color{lamarrOrange}{Euclid}}, {\color{lamarrRed}{Euclid-abs}}$\}$ are dissimilarity measures on attribution scores that we introduce now. 

{\color{lamarrBlue}\textit{Feature Disagreement}} considers only the $k$ top features (indices of $k$ features of highest magnitude) from each of the two explanations and computes then the fraction of common features between them. {\color{lamarrGreen}\textit{Sign Disagreement}} is a more strict version of Feature Disagreement. It applies Feature Disagreement and then subselects only the top features that also have the same sign in both explanations. {\color{lamarrOrange}\textit{Euclid}} and {\color{lamarrRed}\textit{Euclid-abs}} are the Euclidean distance and the Euclidean distance over absolute values of two attribution scores.
Analogously to \cite{krishnaDisagreementProblemExplainable2022}, 
for the disagreement measures we set $k=11$ for AG News, $k=4$ for Dry Bean, and $k=8$ for both Breast Cancer and Ionosphere.

\section{Examining the Rashomon Effect}

We now present and discuss the experiments on numerical stability (Section \ref{sec:111}), the Rashomon Effect itself (Section \ref{sec:011}) and the Disagreement Problem (Section \ref{sec:110}). Each section provides its own discussion.

\label{sec:experiments}

\subsection{Numerical Stability and the Rashomon Effect (111)}
\label{sec:111}
In this section we investigate the setting $\mathop{\mathcal{D}}(f_a, f_a, X=\xref, \phi_1, \phi_1, \text{{\color{lamarrOrange}Euclid}})$ to analyze the numerical stability of all $\phi_*$ w.r.t. differences of individual $f_*$.

Attribution methods often require to choose hyperparameters that control approximation behavior. For IG this is the number of steps used to approximate the integral. For SG, KS, and LI one has control over the number of samples evaluated during computation. This allows to adjust the computation time but if the parameter is too small, the resulting explanations may differ between two computations.
We investigate this approximation stability across many models: Do explanations converge at the same hyperparameter for all models and if not, how large are the differences between individual models?

On the AG News dataset for SG we evaluate sampling hyperparameters $p\in [25, 50, 75,\allowbreak 100, 150]$, for IG, KS, and LI we evaluate $p \in [25, 50, 100, 150, 300]$.
On the tabular datasets we compute the approximation stability for $p \in [25, 50, 75,\allowbreak 100, 125]$ for all methods.
We quantify numerical stability in the following way:
We compute ten SG, KS, and LI explanations for each sample in $\xref$ for each $p$. Next, we compute the pairwise Euclidean distances between all ten explanations. To obtain a stability score for one model, the average is taken across all samples in $\xref$. As a final stability score we report the mean and standard deviation of this score across all models.
IG depends deterministically on the number of steps in the integral, hence, we do not compute ten explanations per sample. Instead, we compute the pairwise distance between explanations for the same point obtained by $p_i$ and $p_{i+1}$. 
To assess model dependent differences regarding the optimal choice of $p$, we compute for each model the smallest $p_i$ in the set of parameters, where $p_{i+1}$ did not improve the average stability by a factor of two.

\begin{table}[htbp]
\caption{Explanation stability for sampling parameter $p$. We report mean$\pm$std across all models and samples for each attribution method. The values for IG describe the difference between using $p_{i+1}$ instead of $p_i$. The last row ($\#$) accumulates the number of models that converged at $\leq p$ for SG/IG/KS/LI.}
    \centering
    \begin{subtable}[t]{\textwidth}
        \centering
            \begin{tabular}[t]{lccccc}
            \toprule[1px]
            & 25 & 50 & 75 & 100 & 150 \\  \midrule
\textbf{SG} & $0.0062 \pm 0.0028$ & $0.0044 \pm 0.0020$ & $0.0036 \pm 0.0016$ & $0.0039 \pm 0.0029$ & $0.0027 \pm 0.0013$ \\ 
            \midrule[1.25px]
            & 25 & 50 & 100 & 150 & 300 \\  \midrule
\textbf{IG} & $0.0734 \pm 0.2740$ & $0.0532 \pm 0.2201$ & $0.0412 \pm 0.1389$ & $0.0329 \pm 0.1380$ & --- \\ 
\textbf{KS} & $6.48e4 \pm 1.3e5$ & $3.34e5 \pm 4.94e5$ & $6.39e6 \pm 1.15e7$ & $1.7121 \pm 0.1655$ & $0.9515 \pm 0.0695$ \\ 
\textbf{LI} & $0.0470 \pm 0.0117$ & $0.0330 \pm 0.0079$ & $0.0220 \pm 0.0055$ & $0.0175 \pm 0.0045$ & $0.0119 \pm 0.0032$ \\ \midrule

            \# & $-/1/-/-$ & $10/2/-/1$ & $18/18/-/19$ & $19/18/20/19$ & $20/20/20/20$ \\

            \bottomrule
            \end{tabular}
        \caption{AG News. Total number of models is 20. The set of evaluated parameters is different from other datasets and different for SG from other methods.}
        \label{tab:111agnews}
    \end{subtable}
    
    \begin{subtable}[t]{\textwidth}
\centering
            \begin{tabular}[t]{lccccc}
            \toprule
            & 25 & 50 & 75 & 100 & 125 \\ \midrule
\textbf{SG} & $0.0005 \pm 0.0003$ & $0.0004 \pm 0.0002$ & $0.0003 \pm 0.0002$ & $0.0003 \pm 0.0001$ & $0.0003 \pm 0.0001$ \\ 
\textbf{IG} & $0.0002 \pm 0.0001$ & $0.0001 \pm 0.0001$ & $0.0001 \pm 0.0000$ & $0.0000 \pm 0.0000$ & --- \\ 
\textbf{KS} & $0.6087 \pm 0.1724$ & $0.2936 \pm 0.0576$ & $0.2232 \pm 0.0422$ & $0.1872 \pm 0.0350$ & $0.1645 \pm 0.0306$ \\ 
\textbf{LI} & $0.1561 \pm 0.0413$ & $0.0956 \pm 0.0272$ & $0.0723 \pm 0.0219$ & $0.0596 \pm 0.0190$ & $0.0515 \pm 0.0170$ \\ \midrule

            \# & $-/-/-/-$ & $39/-/73/96$ & $62/99/99/99$ & $87/99/99/99$ & $99/99/99/99$ \\
            \bottomrule
            \end{tabular}
        \caption{Beans. Total number of models is 99.}
        \label{tab:111beans}
    \end{subtable}
    
    \begin{subtable}[t]{\textwidth}
    \centering
            \begin{tabular}[t]{lccccc}
            \toprule
            & 25 & 50 & 75 & 100 & 125 \\ \midrule

\textbf{SG} & $0.0253 \pm 0.0140$ & $0.0181 \pm 0.0099$ & $0.0148 \pm 0.0081$ & $0.0128 \pm 0.0070$ & $0.0114 \pm 0.0063$ \\ 
\textbf{IG} & $0.0030 \pm 0.0012$ & $0.0013 \pm 0.0006$ & $0.0008 \pm 0.0004$ & $0.0006 \pm 0.0003$ & --- \\ 
\textbf{KS} & $8.28e4 \pm 1.67e5$ & $0.4523 \pm 0.0735$ & $0.3036 \pm 0.0402$ & $0.2462 \pm 0.0312$ & $0.2126 \pm 0.0265$ \\ 
\textbf{LI} & $0.0506 \pm 0.0152$ & $0.0345 \pm 0.0104$ & $0.0279 \pm 0.0085$ & $0.0241 \pm 0.0073$ & $0.0215 \pm 0.0065$ \\ \midrule
            \# & $-/-/-/-$ & $28/-/-/65$ & $58/100/95/86$ & $91/100/100/100$ & $100/100/100/100$ \\
            \bottomrule
            \end{tabular}
        \caption{Breastcancer. Total number of models is 100.}
        \label{tab:111breastcancer}
    \end{subtable}

    \begin{subtable}[t]{\textwidth}
    \centering
            \begin{tabular}[t]{lccccc}

            \toprule
            & 25 & 50 & 75 & 100 & 125 \\ \midrule
\textbf{SG} & $0.0298 \pm 0.0135$ & $0.0209 \pm 0.0095$ & $0.0172 \pm 0.0077$ & $0.0148 \pm 0.0066$ & $0.0133 \pm 0.0060$ \\ 
\textbf{IG} & $0.0036 \pm 0.0017$ & $0.0017 \pm 0.0008$ & $0.0011 \pm 0.0005$ & $0.0009 \pm 0.0004$ & --- \\ 
\textbf{KS} & $1.02e5 \pm 2.06e5$ & $1.81e3 \pm 3.67e3$ & $0.1990 \pm 0.0279$ & $0.1561 \pm 0.0205$ & $0.1325 \pm 0.0169$ \\ 
\textbf{LI} & $0.0995 \pm 0.0239$ & $0.0678 \pm 0.0166$ & $0.0547 \pm 0.0136$ & $0.0472 \pm 0.0119$ & $0.0423 \pm 0.0107$ \\ \midrule
            \# & $-/-/-/-$ & $13/-/-/34$ & $28/51/42/43$ & $47/51/51/50$ & $51/51/51/51$ \\
            \bottomrule

            \end{tabular}
        \caption{Ionosphere. Total number of models is 51.}
        \label{tab:111ionosphere}
    \end{subtable}
    \label{tab:111}
\end{table}
 
Results for all datasets are presented in Table \ref{tab:111}. The rows that start with SG, IG, KS, and LI report numerical stability for each method. 
The last row (\#) reports the accumulated number of models whose explanations are stable at $\leq p$. The aggregated counts correspond to the attribution methods in the order as they appear in the rows: SG, IG, KS, LI. 
Unsurprisingly, numerical stability improves across all models with increasing $p$. 
At the same time, models clearly respond differently to an increase in $p$. 
For SG the spread spans four values of p on each dataset and the selected values for $p$ can differ by a factor of up to three. 
For IG there is no spread on the tabular datasets, but it has the largest spread compared to any other method on AG News. For KS and LI the models mostly split between two consecutive values.
Note that KS displays conspicuously large numerical instability for smaller $p$, even on the smaller tabular datasets. 
Default parameters for KS and LI are set to 25 and 50 in Captum, which is insufficient for a large number of models.
For the remainder of the paper we use explanations computed with the following $p$ for all datasets: SG 100, IG 200, KS and LI 300.

Our results show that, for a rigorous workflow, hyperparameters need to be tuned not only based on the dataset but, in fact, for each model individually. Hence, choosing sensible default parameters is difficult. Providing implementations without default values or with very large values might be an option, though impeding user-friendliness.
This learning also impacts any down-stream use of explanations such as benchmarking methods to assess the fidelity of an attribution method \cite{hooker2019benchmark,alvarezmelis2018,ancona2017towards,yeh2019fidelity,deyoung2019eraser,elshawi2021interpretability} or explanation methods that build atop attributions to extract rules as explanations \cite{vazirgiannisexplaining}. In those contexts, numerical stability is a pre-requisite to obtain reliable results.

\subsection{Solution Diversity  or: The Rashomon effect as seen with Different Dissimilarity Measures (011)}

\label{sec:011}
In this section we investigate how the Rashomon Effect manifests under different metrics over different attribution methods. 
In Table \ref{tab:comparison} we saw that the output behavior of the models is extremely similar.
We are now interested to see how diverse the Rashomon Sets appear if we use the explanation based dissimilarity measure defined above. 
For each dataset and all dissimilarity measures $d \in D$ we evaluate $\mathop{\mathcal{D}}(f_a, f_b, X, \phi_1, \phi_1, d)$ for all pairs of $f_a, f_b \in R$ with $f_a \neq f_b$. Because all attribution scores are specific to the predicted class, we restrict $X \subseteq \xref: \{ x | x \in \xref \land \argmax{f_a(x)} = \argmax{f_b(x)} \}$.

\begin{figure}[ht]
    \centering
    \resizebox{1\textwidth}{!}{\input{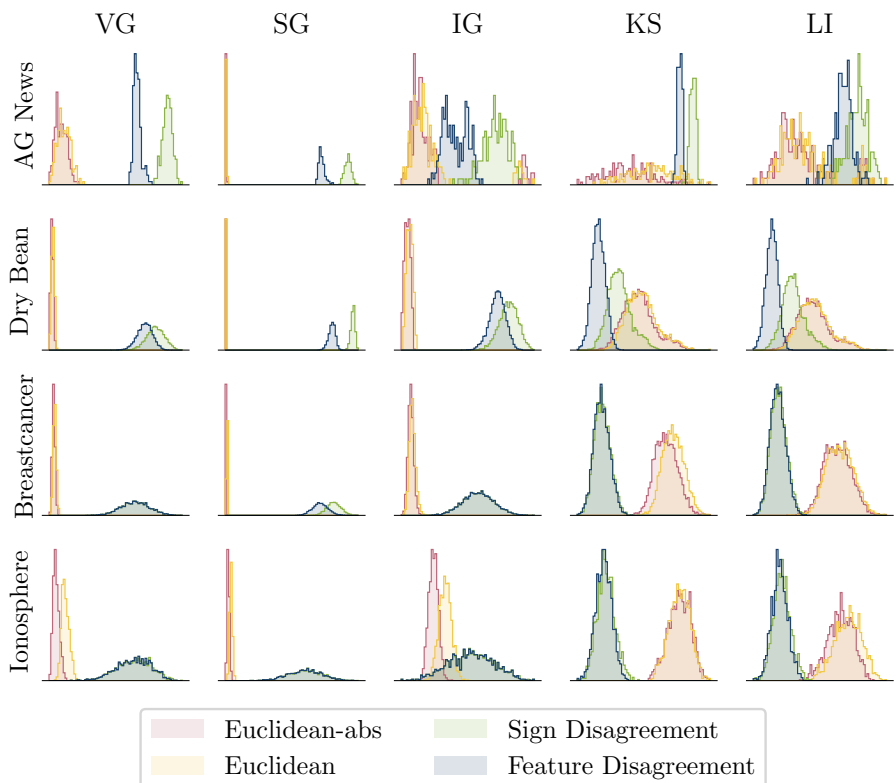}}
    \caption{Histograms over pairwise distances of all models according to Formula \ref{eq:base}. Disagreement metrics computed with $k = 11, 4, 8, 8$ for AG News, Beans, Breast Cancer, and Ionosphere, respectively. 
    In the bottom rows both disagreement metrics overlap nearly exactly.
    }
    \label{fig:disthist}
\end{figure}

The distances produced by Feature Disagreement and Sign Disagreement are naturally bounded to the $[0, 1]$ interval. The gradient based attribution scores lie in a bounded range because we compute the gradient through the softmax output. Attribution scores for KS and LI produced distances larger than 1 with the two Euclidean metrics on all datasets. In those cases we normalize the Euclidean distances to the range $[0, 1]$ by dividing by the maximal distance observed.
Fig. \ref{fig:disthist} visualizes the pairwise distances of all models as histograms.
The x-axis discretizes dissimilarities, the farther to the right the more dissimilar. The y-axis is the number of distances in each bin. 

Euclid and Euclid-abs overlap significantly in all cases except for IG and SG on the Ionosphere dataset. The Disagreement measures diverge on AG News and Dry Bean. Naturally, Sign Disagreement produces larger dissimilarity scores than Feature Disagreement. 

In most of the cases, the means of the disagreement based measures and the Euclidean based measures lie relatively far apart. The exceptions are IG, KS, and LI on AG News, as well as KS and LI on Dry Bean. This means that one metric always measures significantly more differences than the other, but what metric that is depends both on the dataset and method.

We see that with most attribution methods and metrics the Rashomon Sets produce a large variety of distances across all models. This has strong implications for use cases where ML models, specifically (Deep) Neural Networks, are used as epistemic tools to develop hypotheses about the data generation process as it is becoming frequent practice in several disciplines \cite{roscher2020explainable,zednik2022scientific}. 
The variance in our results illustrates that the number of viable solution strategies is extensively large, hence, discovering all possibilities is highly improbable in cases where training a large number of models is infeasible.
Methods such as ROAR \cite{hooker2019benchmark} (despite being developed for a different purpose) could be useful to iteratively narrow down the search space but may still fail to uncover all possible correlations.
The Rashomon Effect also has implications in user-centered scenarios. In cases where users interact with model explanations and expect a certain behavioral consistency over time, the deployment of a new model, even if performance itself is very similar, would pose a risk to user trust.
Depending on the explanation method, the data domain, and the model, computing explanations can be very costly. Storing explanations for later re-use as a way to mitigate costs only works if the model stays the same.

In nearly all cases the human-oriented agreement metrics provide a very different picture than the Euclidean distances. Without additional knowledge about the suitability of a metric in a given context, practitioners should not rely on either disagreement or Euclidean measure alone.
Use cases like \cite{bogun22saliency}, that use explanations to produce training signals for models, could benefit from exploring both kinds of metrics separately or from mixing them in a curriculum.

\subsection{The Rashomon Effect and the Disagreement Problem (110)}
\label{sec:110}

In this section we investigate the Rashomon Effect on the Disagreement Problem. For all datasets and measures $d \in D$ we compare $\mathop{\mathcal{D}}(f_a, f_a, X, \phi_1, \phi_2, d)$ over all pairs $(\phi_1, \phi_2) \in \Phi \times \Phi$ with $\phi_1 \neq \phi_2$. As before, $X$ is the set of all samples in $\xref$ where the predictions of both models agree.

Existing literature on the Disagreement Problem compares disagreement of method pairs for individual or very few models and only with the disagreement measures \cite{neely2021order,krishnaDisagreementProblemExplainable2022,atanasova2020diagnostic,elshawi2021interpretability}. These works report no consistent ranking between method pairs, especially when the data complexity increases.

We now analyze whether we find quantitative support for those observations. Additionally, we extend the analysis of the Disagreement Problem to include results based on the Euclidean distances.

For each individual model we rank the ten possible method pairs from most agreeing to most disagreeing. We calculate Kendall's rank correlation coefficient $\tau$ for all model pairs with a sufficiently small p-value ($< 0.05$). For the remaining $\tau$ the mean and standard deviation across all models are reported in Table \ref{tab:110rkbymodels}.

\begin{table}[t]
    \caption{Kendall rank correlation coefficient ($\tau$) between rankings 
of attribution method pairs. On average we observe a strong or very strong correlation, but the standard deviation indicates that for some models the set of methods that (dis)agree are very different compared to other models.}
    \centering
    \begin{tabular}{lcccc} \toprule
        & AG News & Dry Bean & Breastcancer & Ionosphere\\ 
        \midrule
Feature Disagreement& ${\color{lamarrBlue} 0.67 \pm 0.14 }$ & ${\color{lamarrBlue} 0.65 \pm 0.26 }$ & ${\color{lamarrBlue} 0.66 \pm 0.31 }$ & ${\color{lamarrBlue} 0.63 \pm 0.28 }$ \\  
Sign Disagreement& ${\color{lamarrGreen} 0.63 \pm 0.10 }$ & ${\color{lamarrGreen} 0.57 \pm 0.38 }$ & ${\color{lamarrGreen} 0.64 \pm 0.37 }$ & ${\color{lamarrGreen} 0.73 \pm 0.24 }$ \\ 
Euclidean& ${\color{lamarrOrange} 0.77 \pm 0.29 }$ & ${\color{lamarrOrange} 0.69 \pm 0.15 }$ & ${\color{lamarrOrange} 0.94 \pm 0.12 }$ & ${\color{lamarrOrange} 0.95 \pm 0.11 }$ \\  
Euclidean-abs& ${\color{lamarrRed} 0.79 \pm 0.20 }$ & ${\color{lamarrRed} 0.85 \pm 0.15 }$ & ${\color{lamarrRed} 0.81 \pm 0.18 }$ & ${\color{lamarrRed} 0.84 \pm 0.14 }$ \\  
\bottomrule
    \end{tabular}
    \label{tab:110rkbymodels}
\end{table} 
Two levels of correlation can be observed: 1) Stronger correlation $\gtrsim$ 0.8 for Euclid on AG News, Euclid-abs on Dry Bean as well as both Euclidean based metrics on Breast Cancer and Ionosphere. 2) A moderate correlation $\approx 0.65$ for Feature Disagreement on all datasets with a lower standard deviation on AG News.
Sign Disagreement also falls in this range on all datasets but Dry Bean, with a notably lower standard deviation on AG News compared to other datasets. 
The lowest correlation (0.57) is produced by Sign Disagreement on Dry Bean, showing the largest standard deviation (0.38) at the same time.
The large standard deviations suggest that a fair amount of models produces very different rankings, particularly in the case of the disagreement based rankings.

\begin{figure}[t]
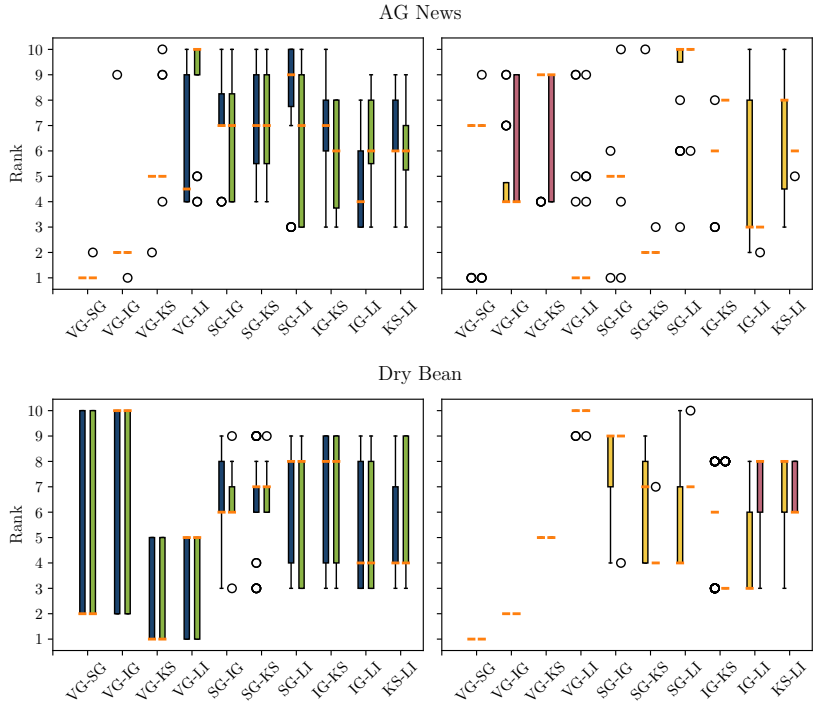

    \centering
    \begin{subfigure}{.9\textwidth}
        \resizebox{1.\textwidth}{!}{\input{figures/agnews/agnews_110_rk_bp.pgf}}
\end{subfigure}
\begin{subfigure}{.9\textwidth}
        \resizebox{1.\textwidth}{!}{\input{figures/beans/beans_110_rk_bp.pgf}}
\end{subfigure}
    \caption{Box plots of rankings over which pair of attribution methods disagrees most or least for individual models on AG News (top) and Dry Bean (bottom). Higher rank means larger disagreement. Plots on the left: {\color{lamarrBlue} Feature Disagreement} {\color{lamarrGreen} Sign Disagreement}, plots on the right: {\color{lamarrRed} Euclid-abs} {\color{lamarrOrange} Euclid}; Orange lines in each boxplot indicate the median.}
    \label{fig:110rk_agnews_beans}
\end{figure}

\begin{figure}[t]
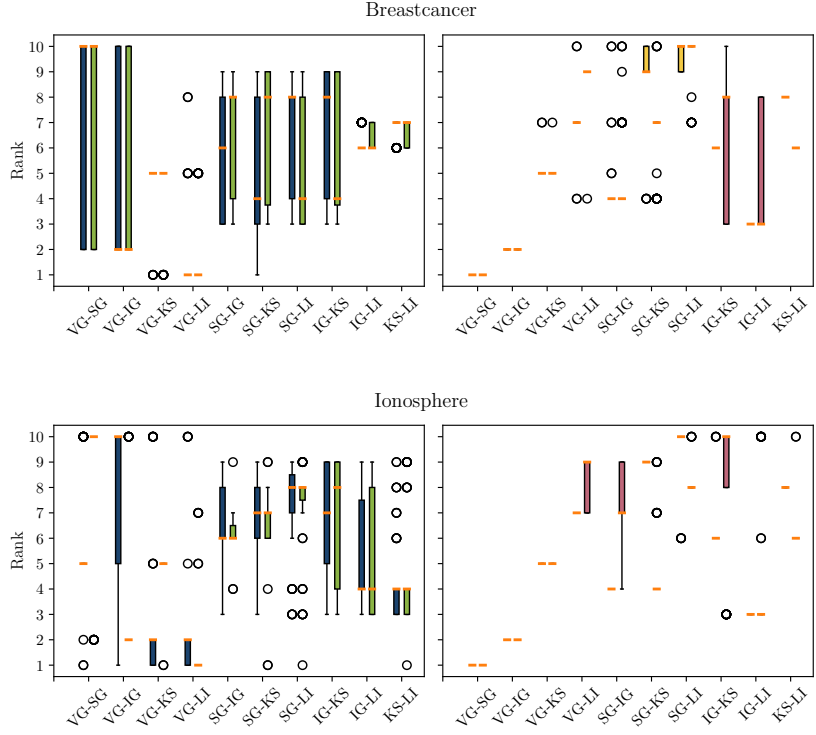

    \centering
    \begin{subfigure}{.9\textwidth}
        \resizebox{1\textwidth}{!}{\input{figures/breastcancer/breastcancer_110_rk_bp.pgf}}
\label{fig:110rk_bc}
    \end{subfigure}
    \begin{subfigure}{.9\textwidth}
        \resizebox{1\textwidth}{!}{\input{figures/ionosphere/ionosphere_110_rk_bp.pgf}}
\label{fig:110rk_iono}
    \end{subfigure}
    \caption{Box plots of rankings over which pair of attribution methods disagrees most or least for individual models on Breast Cancer (top) and Ionosphere (bottom). Higher rank means larger disagreement. Plots on the left: {\color{lamarrBlue} Feature Disagreement} {\color{lamarrGreen} Sign Disagreement}, plots on the right: {\color{lamarrRed} Euclid-abs} {\color{lamarrOrange} Euclid}; Orange lines in each boxplot indicate the median.}
    \label{fig:110_rk_bc_iono}
\end{figure}

Are lower correlations structural? I.e. is it always specific method pairs that tend to swap ranks?
We visualize the rank that each pairing occupies for every model in the box plots in Fig.~\ref{fig:110rk_agnews_beans} and Fig.~\ref{fig:110_rk_bc_iono}.
The y-axis shows the rank, higher rank meaning stronger disagreement relative to the other methods. Plots on the left pair both disagreement based rankings ({\color{lamarrBlue}blue}/ {\color{lamarrGreen}green}) while plots on the right show results for Euclidean based rankings ({\color{lamarrOrange}yellow}/ {\color{lamarrRed}red}).

Generally, we can make the following observations about the Euclidean metrics: 1) For most explanation pairs, both Euclidean metrics show little to no variance within each dataset, signifying agreement on the ranking of the respective explainability pair, but no consistent ranking across all datasets. 2) All three tabular datasets agree for VG-SG being on rank one and VG-IG being on rank two, both with no variance.

Looking at the results for the disagreement metrics for each dataset in detail, we can see the following:
AG News (Fig. \ref{fig:110rk_agnews_beans}) shows very stable rankings for both Disagreement metrics for VG-\{SG, IG, KS\}. 
For Dry Bean (Fig. \ref{fig:110rk_agnews_beans}) across both disagreement metrics, the median lies 8/20 times exactly on one of the quartiles which show no whisker.
This means that 50\% of the models agree on the respective ranking. 
This is interesting because at the same time VG-\{SG, IG\} span nearly the whole ranking, meaning that all rankings in the fourth quartile assign the maximum rank. The pairs SG-\{IG, KS\} seem to swap places but are otherwise rather consistently placed in the lower middle of the ranking. 
Breast Cancer (Fig. \ref{fig:110_rk_bc_iono}) shows stable rankings for VG-\{KS, LI\} with both disagreement metrics. More interestingly, for VG-\{SG, IG\} the medians lie again on ``whiskerless''-quartiles and the ranking agrees with the one on Dry Bean (rank 2 for VG-SG and rank 10 for VG-IG). In contrast to Dry Bean, here it is IG-LI and KS-LI that place comparably stable towards the middle of the ranking.
On Ionosphere (Fig. \ref{fig:110_rk_bc_iono} bottom) the plot shows smaller boxes compared to the other tasks. Taking outliers into account, multiple pairings span the whole ranking for disagreement based rankings. Ignoring outliers, there are five stable rankings for VG-\{IG, SG, KS, LI\}, four of which are achieved with Sign Disagreement. 

We summarize our observations: 
We did not see a consistent ranking across all datasets and metrics. 
Our results for the disagreement based metrics support the observation from the literature that there is no consistent ranking among method pairs. However, we do not observe that results on the more complex AG News appear less correlated than for smaller tabular tasks.
Our evaluation of Euclidean based rankings shows them to be notably more stable than their disagreement counterparts.

Interestingly, we cannot identify a single pair of methods that produces high disagreement across all tasks and metrics consistently, but there are pairs of methods for each dataset that consistently take mid-range rankings.
Practitioners that seek diverse explanations would be recommended to start their search with comparing VG-KS, SG-\{IG, KS\}, and KS-LI.

\subsection{Summary}
\label{sec:4}

In the first scenario in Section \ref{sec:111} we evaluated how sensitive individual models are to hyperparameter choices for non-deterministic attribution methods. Expectedly, a higher sampling rate always improves the numerical stability of the approximations. However, we found stark differences between the individual models, some requiring larger parameter values by a factor of up to twelve. This has direct implications for scientists and developers using XML methods, as it means that prior knowledge is not necessarily transferable between two models. Choosing default values is de facto impossible. Especially scenarios where parameters have to be chosen as small as possible require rigorous testing.

After verifying the numerical stability of our explanations, in Section \ref{sec:011} we assessed how the Rashomon Effect manifests itself on different datasets, depending on the different attribution methods and dissimilarity measures. We illustrated the solution diversity under different dissimilarity measures. We found that gradient based attribution methods in conjunction with Euclidean metrics showed smaller distances and low variance on the simpler tabular datasets. 
Disagreement based dissimilarity measures produced high distances and variances in nearly all cases. The distances are notably higher for Sign Disagreement compared to Feature Disagreement in half of the cases.
We saw a large spectrum of distances for perturbation based methods in all cases.
Our observation of large magnitudes and high variances in the distances has implications for ML as an epistemic tool. 
It illustrates how large the space of possible viable solution strategies is, indicating the need to develop informed search strategies in the future \cite{beckh2023}, especially in complex or resource constrained scenarios.
Also, the histograms of Euclidean and disagreement-based measures rarely show overlap, meaning practitioners will have to make context-specific choices on what type of metric to use.
In cases where model behavior is explained to users, deploying a model update can lead to irritations as the explanations will likely change drastically between any two models; using the computationally intensive KS or LI seems to give the best chances to maintain somewhat consistent explanations.
Conversely to the use-case of ML as an epistemic tool, a possible direction of future work is the inverse search problem of finding a better performing model that functions most similarly.

Investigating the Rashomon Effect on the Disagreement Problem in Section \ref{sec:110} revealed stark differences between results from the disagreement measures and the Euclidean distances. Neither of the two metric types produced rankings consistent across all datasets. 
Within tasks, the Euclidean metrics produced very stable rankings while 
the disagreement measures only occasionally produced a stable rank for a few pairs. Thus, our work provides quantitative support to the observations in  \cite{elshawi2021interpretability,krishnaDisagreementProblemExplainable2022,neely2021order} that are based on a small number of models, only. However, contrary to the literature, our results do not look more stable for smaller models on tabular datasets than for the Bi-LSTM model on AG News.

\section{Conclusion}
\label{sec:5}
We have quantitatively shown how the Rashomon Effect impacts the application and interpretation of XML techniques and argue that it has to be taken into account by the XML community in the future.
Along the three variables 1) the model, 2) the datum, and 3) the attribution method, we presented a structured investigation of the Rashomon Effect from three perspectives within XML. 

Our quantitative analysis on numerical stability showed models to have individual sensitivity to hyperparameters of explanation methods. We have shown that choosing the most efficient setting requires careful tuning not only to a specific task or architecture, but in fact to every model instance individually, in order to guarantee stable explanations. 
Assessing the Rashomon Effect itself by measuring the diversity of solution strategies, we found that the solution space appears extensive, especially under the disagreement metrics. This poses challenges to applications of ML as an epistemic tool, as well as use cases where models are offered to consumers that expect consistent behavior.
Our study of the Disagreement Problem provides quantitative support for previously anecdotal evidence. No consistent ranking persists across all datasets and the only option for practitioners that seek diverse explanations is trial and error. However, for each dataset individually we were able to identify a pair of methods that consistently take mid-range ranks. 
Using those rankings to systematically compare methods might yield insight into differences regarding what parts of model behavior each method is sensitive to.

\subsubsection*{Acknowledgments}
This research has been funded by the Federal Ministry of Education and Research of Germany and the state of North-Rhine Westphalia as part of the Lamarr-Institute for Machine Learning and Artificial Intelligence Lamarr22B.
Part of PWs work has been funded by the Vienna Science and Technology Fund (WWTF) project ICT22-059.

\section*{Ethical Statement}

In critical contexts, where persons are directly or indirectly impacted by a model, and where explanations are used to verify that model behavior is compliant with a given standard, proper use of explanation methods is of utmost importance.
Hyperparameter choices have to be validated for each model individually. For model testing and validation procedures to be reliable they have to integrate this knowledge.
Our work demonstrated that it is unreasonable to expect an explanation computed for one model, to be valid for another model, however similar their performance otherwise may be. 
Re-using explanations from one model to give as an explanation of behavior for another model is not possible and has to be avoided in critical scenarios.

\bibliographystyle{splncs04}
\bibliography{literature}

\end{document}